%% file: main.tex
\documentclass[table]{article} 
\usepackage{iclr2024_conference,times}

\input{math_commands.tex}

\usepackage{hyperref}
\usepackage{url}
\usepackage{multirow}
\usepackage{booktabs}
\usepackage{graphicx}
\usepackage{siunitx}  
\usepackage{tabularx}
\usepackage{highlight}
\usepackage[flushleft]{threeparttable}
\usepackage{enumitem}
\usepackage{amsmath}
\usepackage{xspace}
\usepackage{cleveref}
\usepackage{caption}
\usepackage{subcaption}
\usepackage{pgfplots}
\usepackage{tikz}
\usepackage{wrapfig}
\usepackage{authblk}
\usepackage{fdsymbol}
\usepackage{listings}
\definecolor{mymauve}{rgb}{0.58,0,0.82}
\newcommand{\lemuR}[0]{Lemur\xspace}
\newcommand{\lemurSeventyB}[0]{\texttt{Lemur-70B}\xspace}
\newcommand{\lemurChaT}[0]{Lemur-Chat\xspace}
\newcommand{\lemurSeventyChaT}[0]{\texttt{Lemur-70B-Chat}\xspace}

\newcommand{\llamaTwO}[0]{Llama-2\xspace}

\newcommand{\llamaTwoSeventY}[0]{\texttt{Llama-2-70B}\xspace}
\newcommand{\llamaTwoSeventyChaT}[0]{\texttt{Llama-2-70B-Chat}\xspace}
\newcommand{\starcoderFiveteenB}[0]{\texttt{StarCoder-15B}\xspace}
\newcommand{\starcoderPlusFiveteenB}[0]{\texttt{StarCoderPlus-15B}\xspace}
\newcommand{\wizardcoderFiveteenB}[0]{\texttt{WizardCoder-15B}\xspace}

\newcommand{\codellamaThirtyFouR}[0]{\texttt{CodeLlama-34B}\xspace}
\newcommand{\codellamaThirtyFourBInstuctioN}[0]{\texttt{CodeLlama-34B-INST}\xspace}
\newcommand{\gptturbo}[0]{\texttt{gpt-3.5-turbo}\xspace}
\newcommand{\gptfour}[0]{\texttt{gpt-4}\xspace}
\definecolor{rebuttal}{HTML}{000000}

\title{
\textsc{Lemur}: Harmonizing Natural Language and Code for Language Agents}

\author{
        \textbf{Yiheng Xu}\thanks{Equal contribution.}~~$^\spadesuit$$^\diamondsuit$\ \
        \textbf{Hongjin Su}$^*$$^\spadesuit$$^\diamondsuit$\ \ 
        \textbf{Chen Xing}$^*$$^\clubsuit$\ \
        \textbf{Boyu Mi}$^\spadesuit$$^\diamondsuit$\ \ 
        \textbf{Qian Liu}$^\heartsuit$$^\diamondsuit$\ \
        \textbf{Weijia Shi}$^\triangle$\\
        \vspace{-10pt}
        \textbf{Binyuan Hui}$^\diamondsuit$
        \textbf{Fan Zhou}$^\spadesuit$$^\diamondsuit$
        \textbf{Yitao Liu}$^\spadesuit$$^\diamondsuit$ 
        \textbf{Tianbao Xie}$^\spadesuit$$^\diamondsuit$
        \textbf{Zhoujun Cheng}$^\spadesuit$$^\diamondsuit$
        \textbf{Siheng Zhao}$^\spadesuit$$^\diamondsuit$
        \textbf{Lingpeng Kong}$^\spadesuit$ 
        \textbf{Bailin Wang}$^\bigstar$ 
        \textbf{Caiming Xiong}$^\clubsuit$ 
        \textbf{Tao Yu}$^\spadesuit$$^\diamondsuit$ \\
  $^\spadesuit$University of Hong Kong
  \quad
  $^\diamondsuit$XLang Lab
  \quad
  $^\clubsuit$Salesforce Research \\
  $^\heartsuit$Sea AI Lab \ \
  \quad
  $^\triangle$University of Washington
  $^\bigstar$MIT CSAIL \\
  $^\spadesuit${\texttt{\{yhxu,hjsu,tyu\}@cs.hku.hk}}\ \
  $^\clubsuit${\texttt{\{cxing, cxiong\}@salesforce.com}}
}

\iclrfinalcopy 

\begin{document}

\newtheorem{exmp}{Example}[section]
\maketitle
\setlength{\abovedisplayskip}{2pt}
\setlength{\belowdisplayskip}{2pt}

\begin{abstract}
We introduce Lemur and Lemur-Chat, openly accessible language models optimized for both natural language and coding capabilities to serve as the backbone of versatile language agents. 
The evolution from language chat models to functional language agents demands that models not only master human interaction, reasoning, and planning but also ensure grounding in the relevant environments.
This calls for a harmonious blend of language and coding capabilities in the models.
\lemuR and \lemurChaT are proposed to address this necessity, demonstrating balanced proficiencies in both domains, unlike existing open-source models that tend to specialize in either.
Through meticulous pre-training using a code-intensive corpus and instruction fine-tuning on text and code data, our models achieve state-of-the-art averaged performance across diverse text and coding benchmarks. 
Comprehensive experiments demonstrate \lemuR's superiority over existing open-source models and its proficiency across various agent tasks involving human communication, tool usage, and interaction under fully- and partially- observable environments.
The harmonization between natural and programming languages enables Lemur-Chat to significantly narrow the gap with proprietary models on agent abilities, providing key insights into developing advanced open-source agents adept at reasoning, planning, and operating seamlessly across environments. 
Our model and code have been open-sourced at \url{https://github.com/OpenLemur/Lemur}.
\end{abstract}

\section{Introduction}
Intelligent agents are broadly conceptualized as autonomous problem solvers with the abilities to sense their environment, decide, and act upon that enviorment~\citep{Sutton2005ReinforcementLA, russell2010artificial, wilkins2014practical}.
Recent implementations of this concept in creating language agents~\citep{yao2022react, AutoGPT, wang2023voyager} capable of utilizing natural language for varied and intricate tasks in diverse environments have demonstrated potential, particularly when built upon large language models (LLMs)~\citep{brown2020language, chen2021evaluating, chowdhery2022palm, openaiGPT4TechnicalReport2023, touvronLlamaOpenFoundation2023}.
Such agents harness the human knowledge in LLMs and can think and communicate in human terms. 
This equips them to employ varied tools, operate in complex environments, engage in language reasoning, and create spontaneous multi-agent systems.

To effectively form the foundation of language agents, LLMs should not only master human interaction, reasoning, and planning but also ensure grounding in the relevant environments~\citep{wei2022chain, huang2022language, ahn2022can}.
Human interaction, reasoning, and planning can be largely realized through the natural language capabilities of LLMs.
On the other hand, the grounded execution in the environment is usually achieved by using general-purpose code or domain-specific APIs, such as controlling web browsers~\citep{shi2017world, yao2022webshop, deng2023mind2web, zhou2023webarena}, interacting with OS CLI terminals~\citep{yang2023intercode}, and manipulating robotic arms via primitive APIs~\citep{ahn2022can, huangInnerMonologueEmbodied2022, liang2023code}. 
Therefore, we posit that for the construction of language agents, it is imperative for language models to possess harmonized capabilities in both natural language and programming languages. 
This balance ensures that models do not specialize exclusively in certain areas but can seamlessly integrate with environment contexts and generate controllable and valid actions.
Presently, closed-source models like GPT-4~\citep{openaiGPT4TechnicalReport2023} demonstrate such capabilities, which empower them to function as language agents. 
However, current open-source LLMs such as Llama 2~\citep{touvronLlamaOpenFoundation2023} and CodeLlama~\citep{roziereCodeLlamaOpen2023} have traditionally been tailored for either textual or code-related tasks, with limited ability to effectively balance both.

To address this need, we introduce \lemuR and \lemurChaT, cutting-edge, openly accessible models pre-trained and fine-tuned to harmonize text and code capabilities. 
We enhanced the base Llama-2-70B through thoughtfully designed pre-training and instruction fine-tuning stages.
Specifically, we built a code-centric corpus based on The Stack~\citep{kocetkovStackTBPermissively2022}, comprising 90B tokens with a 10:1 code-to-text ratio, ensuring improved capabilities in coding ability while maintaining performance in natural language ability.
We refer to this model as \lemuR. 
After pretraining, we conducted instruction fine-tuning using about 300K examples from both text and code to build an instruction-following model, which we refer to as \lemurChaT.
Thorough assessments across 8 textual and coding benchmarks validate the superior performance of both Lemur and Lemur-Chat in multiple text and code evaluations, establishing them as the most well-rounded open-source models.

Moreover, this work embarks on assessing the vital capabilities of language agents across various scenarios, which we refer to as agent benchmarks.
We place a particular emphasis on their tool-usage abilities, and abilities in grounding in the environment feedback and human feedback.
We also explore the challenges posed by real and partially observable environments where the agent has to take actions based on limited knowledge and take additional actions to gather more information.
Experimental results indicate that \lemurChaT outperforms other open-sourced models in 12 of the 13 agent benchmarks. 
This underscores how the integration of natural and coding capabilities allows \lemurChaT to exceed the current open-source models for language agents, markedly bridging the performance disparity between open-source and commercial alternatives.
Our experiments show {\color{rebuttal} an important insight} that in agent scenarios, there is a need for synergy between natural language and coding abilities. Specifically, for models with strong natural language capabilities but weak coding abilities like \llamaTwoSeventyChaT, they can effectively use simple tools to assist reasoning (\S\ref{sec:tools}) because the action space is small and the difficulty of using tools is low. However, when facing complex decision-making scenarios such as web browsing and house navigation, the action space is usually large, and models with strong coding abilities have an advantage in generating complex executable action sequences (\S\ref{sec:partiallyobservable}).
{\color{rebuttal} Moreover, we did error analysis on several environments covering multiple agent skills to reflect the key challenges for future language agent development, including tool and action executability~\S\ref{appendix:error_analysis_math}, problem difficulty~\S\ref{appendix:error_analysis_sql}, and domain knowledge~\S\ref{appendix:error_analysis_ctf}.}
Overall, Lemur has both strong natural language and coding abilities, enabling it to achieve better performance in both scenarios. This research provides insights into optimizing the synergy between natural and programming languages, laying a solid foundation for developing advanced language agents capable of operating efficiently in various environments.

\vspace{-5pt}
\section{Pre-training and Instruction tuning of \lemuR}
\begin{figure}[ht]
    \centering
    \includegraphics[width=0.95\textwidth]{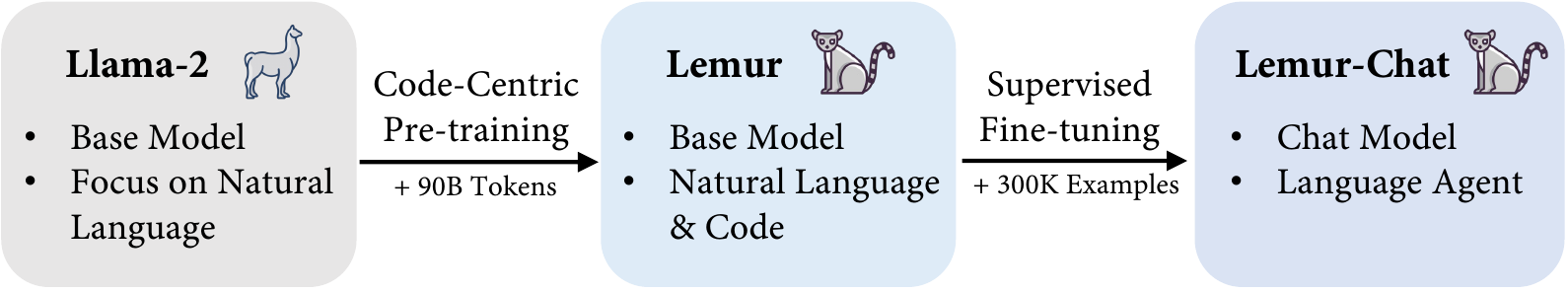}
    \caption{Overview of Training Procedure. We continually pre-train \llamaTwoSeventY model on 90B code-intensive data and fine-tune it with 300K examples of instructions to enable the model to harmonize natural language and coding abilities.}
    \label{fig:train}
\end{figure}

This section will introduce the method to build the \lemuR and \lemurChaT models and their performance on commonly-used benchmarks for pre-trained language model evaluation.
To build a more balanced model, the pipeline includes two stages: pre-training~(\S~\ref{sec:pre-training}) and instruction fine-tuning~(\S~\ref{sec:instruction-finetuning}). 
The training pipeline is shown in \Cref{fig:train}.

\subsection{Pre-training}
\label{sec:pre-training}
In the pre-training stage, we choose the \llamaTwoSeventY base model as the starting point, which is the cutting-edge open-sourced base model in most scenarios except the coding scenario. 
Our goal is to improve the coding ability while maintaining the reasoning ability of \llamaTwO. 
To this end, we built a corpus with a code-to-text ratio of 10:1. 
{\color{rebuttal} We discuss how we decided this ratio in \S~\ref{appendix:ratio}.}
For the code part, we base it on The Stack~\citep{kocetkovStackTBPermissively2022}, a collection of source codes from GitHub with permissive licenses. 
Among all languages, we focus on scripting or interpreted languages (Python, SQL, Bash, Perl, etc.) because agent models are often executed in interactive scenarios.
{\color{rebuttal} Unlike compiled languages like C++, the interpreted nature of scripting languages allows for immediate execution and ease of modification, which is essential for dynamic interaction in language agent scenarios. }
As for the text aspect, we use RefinedWeb~\citep{penedoRefinedWebDatasetFalcon2023}, Redpajama~\citep{together2023redpajama}, as well as CommonCrawl, Wikipedia, Books, ArXiv, StackExchange and DM Mathematics~\citep{saxtonAnalysingMathematicalReasoning2019} to build the textual data corpus. 
Following previous works~\citep{gaoPile800GBDataset2020a, smithUsingDeepSpeedMegatron2022, together2023redpajama, liStarCoderMaySource2023}, we do extensive deduplication after aggregating all data sources. 
The composition of the data is shown in Appendix~\S~\ref{appendix:ptdata}. 
We train the \lemurSeventyB model initialized with \llamaTwoSeventY using a TPUv4-512 pod. 
We train the model with sequence packing~\citep{raffelExploringLimitsTransfer2019, chungScalingInstructionFinetunedLanguage2022} to improve the training efficiency.
Please refer to Appendix \S~\ref{appendix:pttraining} for more details.

\subsection{Instruction Fine-tuning}
\label{sec:instruction-finetuning}
During the instruction fine-tuning phase, we include four data sources to construct \lemurChaT, including the Open Assistant crowdsourced annotated dialogue corpus~\citep{kopfOpenAssistantConversationsDemocratizing2023}, Orca data with chain of thought reasoning for human-written tasks~\citep{OpenOrca, mukherjeeOrcaProgressiveLearning2023}, ShareGPT \& Chatlogs containing real user and ChatGPT history records~\citep{sharegpt_data}, as well as Evol-CodeAlpaca data~\citep{luo2023wizardcoder} consisting of complex coding tasks generated by ChatGPT along with their solutions. 
The statistics of these instruction datasets are shown in \Cref{tab:instruction-tuning-datasets}.
After we collect these data, we additionally clean and deduplicate these instruction fine-tuning data. 
We conduct training on these data for 2 epochs.
Please refer to \S~\ref{appdx:ift} for more details.
\input{tables/sftdata}

\section{From Language Model to Language Agent}
This section introduces how we measure the language and coding abilities of a language model to guide the process of harmonization. 
We further discuss the new challenges faced when connecting LLM to the environment and describe how we examine the necessary agent capabilities.

\subsection{Fundamental Langauge and Coding Capabilities}
\label{sec:static_evaluation}

Previous work typically uses a variety of benchmarks to comprehensively reflect the performance of models on different types of tasks. 
Therefore, we evaluate the performance of various models across text and code benchmarks as follows. 
\begin{itemize}[leftmargin=0.1 in]
\item \textit{Text benchmarks:} MMLU~\citep{hendrycks2021measuring} to determine factuality, BBH~\citep{suzgun2022challenging} to check reasoning abilities, GSM8K~\citep{cobbe2021training} to gauge math reasoning.
\item \textit{Code benchmarks:} HumanEval~\citep{chen2021evaluating} and MBPP~\citep{austin2021program} to test Python writing abilities, Spider~\citep{yu2018spider} to assess database query capabilities, MultiPL-E~\citep{10103177} to measure the multi-lingual coding capabilities, DS-1000~\citep{lai2023ds} for evaluation in data science scenarios
\end{itemize}

\subsection{Connecting LLM Agents to Environment}
\begin{figure}[ht]
    \centering
    \includegraphics[width=1.0\textwidth]{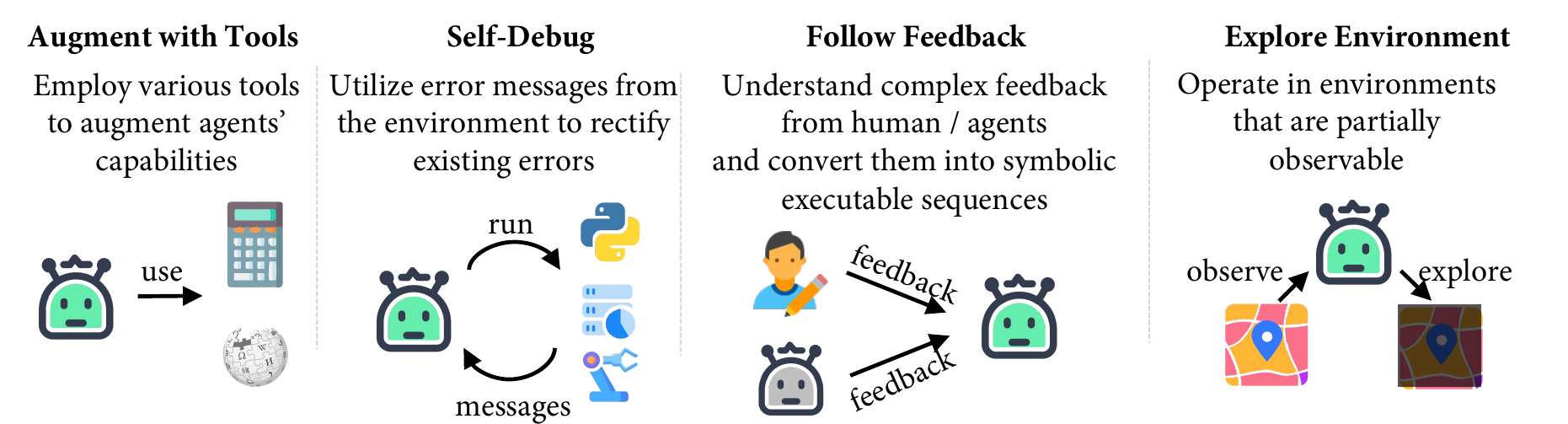}
    \caption{We inspect language agents in various aspects, including the abilities to augment with tools, self-debug, follow feedback, and explore partially observable environments.}
    \label{fig:agent}
\end{figure}

While the measurements in \S\ref{sec:static_evaluation} provide valuable insights on each domain, they may not fully encapsulate the models' capabilities in language agent scenarios due to their focus on single-turn interactions in fully observable settings.
In order to address this discrepancy, we scrutinize models in the context of multi-turn interactive scenarios to gauge their adaptability in practical situations. 
Our assessment is centered on various capabilities of language agents, which are encapsulated in the factors outlined in \Cref{fig:agent} and \Cref{tab:eval_datasets}. 
For a more comprehensive assessment of these capabilities, we reorganize existing multiple datasets into four sets to examine the diverse skills demanded by the agent scenario. 
Please refer to Appendix~\ref{appendix:agent-eval} for details of each dataset.

\input{tables/agent-datasets}

\textbf{Augment with Tools}
Tool usage~\citep{schick2023toolformer, haoToolkenGPTAugmentingFrozen2023, mialonAugmentedLanguageModels2023} is an important capability of language agents. Tasks like calculations or information retrieval can be offloaded to external modules (e.g. Python interpreters or search engines) using tools~\citep{gao2023pal, chen2022program}, improving reliability and interpretability. Tools involve calling operators in symbolic languages or invoking APIs~\citep{shen2023hugginggpt}. This requires decomposing a task, grounding intention to tools, and using the results for further actions, relying on natural language reasoning and programming abilities~\citep{Binder, viper2023}.
To assess the ability of language agents to solve complex multi-turn problems using tools, we introduce the part of the MINT dataset~\citep{wang2023mint} which focuses on tool-utilization for reasoning. 
This part includes several adapted datasets for testing: MINT-\{GSM8K, MATH\} ~\citep{cobbe2021training,math2022dataset} is used to test the model’s ability to solve mathematical problems using a Python interpreter, and MINT-\{HotpotQA, MMLU\}~\citep{yang2018hotpotqa,hendrycks2021measuring} assesses the model's capability to solve knowledge-based questions using Wikipedia searches. 
At the same time, for the MINT-TheoremQA~\citep{chen2023theoremqa}, the model needs to perform knowledge searches on Wikipedia and use a Python interpreter to calculate and draw conclusions.

\textbf{Self-debug with Environment Feedback}
Self-debug is an important way to test whether a model can incorporate environmental feedback~\citep{jignasu2023towards,olausson2023demystifying,chen2023teaching}. 
In the self-debug scenario, the model usually needs to complete a complex operation sequence, such as Python functions, database queries/modifications, robot action sequences, etc~\citep{gur2023real,yao2023retroformer}. These complex operations sometimes cannot be executed successfully and will return error messages. 
This requires the model to comprehend this kind of environmental feedback and correct errors, which tests the joint effect of natural language reasoning ability and programming languages. 
We use rich datasets from multiple benchmarks to evaluate this performance, including multi-turn MINT-MBPP and MINT-HumanEval in MINT~\citep{wang2023mint}, SQL and Bash in InterCode~\citep{yang2023intercode}, as well as RoboCodeGen that calls robot APIs through code~\citep{liang2023code}. 
These environments require the model to complete complex tasks and will provide execution errors. In these environments, model performance will vary according to its self-debugging ability, reflecting the ability to incorporate feedback.

\textbf{Follow Natural Language Feedback}
Following natural language feedback is an important mechanism for agents to receive information from humans or other agents~\citep{wang2022self,ouyang2022training,gong2023mindagent}. 
In scenarios where complex problems are solved through multi-turn interactions, the model not only needs to incorporate environmental feedback but also feedback from humans or other agents in order to improve. 
This mechanism requires the model to understand new instructions based on a context that combines natural language and code, and ground them into new action sequences. 
To evaluate the model's ability to accept natural language feedback, we follow the approach of the MINT benchmarks: using a GPT-4 simulated user as a teacher to guide the model in problem-solving. 
This setup includes a series of MINT datasets~\citep{wang2023mint} to comprehensively evaluate performance after adding natural language feedback in various scenarios.

\textbf{Explore in Partially Observable Environments}
Exploring partially observable environments is a unique and challenging factor in agent scenarios. 
All the settings mentioned earlier can be considered as fully observable environments, which means that agents can observe all the information of the environment to plan, reason, and make decisions. 
However, in partially observable environments (also known as Partially Observable Markov Decision Process)~\citep{kurniawati2021partially}, agents can only partially observe the environmental information to solve problems. 
This requires agents to collect information through exploration and continue making decisions. 
This process places high demands on various abilities of agents, such as natural language planning and reasoning, environmental interaction, etc. 
To measure this ability, we use three datasets: InterCode-CTF~\citep{yang2023intercode} and WebArena~\citep{zhou2023webarena} in digital environments, as well as ALFWorld~\citep{shridhar2020alfworld} in physical environments. 
InterCode-CTF provides an OS terminal for models to solve Catch the Flag (CTF) problems where agents need multiple rounds of exploration to obtain the flag. 
WebArena evaluates agents' ability to control browsers for task completion through exploration. 
ALFWorld is a simulated home environment where agents need to explore navigation and complete specific tasks.

\section{Experimental Results}

\subsection{Language and Code Capabilities}
The comprehensive evaluations of text and code benchmarks in \Cref{tab:base&instruct} demonstrate the impressive capabilities of \lemurSeventyB and \lemurSeventyChaT models. 
Deviating from \llamaTwoSeventY and \llamaTwoSeventyChaT, which are mostly pre-trained and fine-tuned on text data, the \lemuR models augment coding abilities and thereby enhance the overall performance by 4.3\% and 14.8\% respectively. 
Alternatively, models like \color{rebuttal}{StarCoder-15B,  WizardCoder-15B, \codellamaThirtyFouR and \codellamaThirtyFourBInstuctioN, which are primarily trained on code datasets, exhibit solid performance in code benchmarks.
\color{rebuttal}{On average, \lemurSeventyB outperforms StarCoder-15B, StarCoderPlus-15B and \codellamaThirtyFouR} by 14.3\%, 16.8\% and 1.9\% respectively, and \lemurSeventyChaT surpasses WizardCoder-15B and \codellamaThirtyFourBInstuctioN by 18.0\% and 9.4\% respectively. 
This commendable increase highlights the virtue of harmonizing textual and coding skills.}

\color{black}
The synergic text and code abilities enable them to function as language agents.
However, a disparity exists between natural language and coding capabilities in current open-source models. 
Such limitations obstruct these models' abilities to act as language agents, leading to performance degradation in agent benchmarks.
In subsequent sections, we meticulously evaluate various critical capabilities of agents, revealing the importance of synergic text and code abilities to language models.

\input{tables/tab1-base-instruct}

\subsection{Augment with Tools}
\label{sec:tools}
In the realm of problem-solving, agents, akin to humans, employ various tools to augment their capabilities. 
This is exemplified by~\citep{chen2022program}, who showcase that the mathematical reasoning prowess of LLM can be significantly enhanced with the aid of Python. 
As per the data presented in Table~\ref{tab:tooluse}, it is evident that \lemurSeventyChaT outperforms both \llamaTwoSeventyChaT and \codellamaThirtyFourBInstuctioN, indicating its superior ability to effectively leverage tools.
{\color{rebuttal}In addition, we found performance differences between open-source models and closed-source models in the challenging math benchmark tests M-MATH and M-TheoremQA. We further analyzed the errors to better understand this phenomenon, please refer to the content in Appendix \ref{appendix:error_analysis_math}.}

\setlength\intextsep{0pt}
\begin{wrapfigure}[15]{R}{0.45\textwidth}
     \centering
\input{images/intercode-line.tex}
\vspace{-4mm}
\captionsetup{margin={0.6cm,0cm}}
         \caption{Success Rate with interaction turns across models in IC-SQL.}
         \label{fig:intercoder-sql}
\end{wrapfigure}
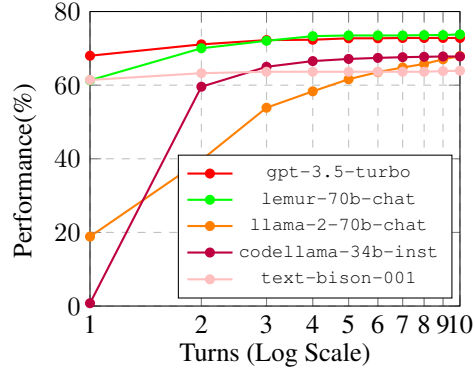

\vspace{3pt}
\input{tables/tab2-tooluse}

\subsection{Self-debug with Environment Feedback}
\label{sec:self-debug}
The technique of self-debug gains considerable traction in the realm of code generation~\citep{olausson2023demystifying,zhang2023self}. 
This method consists of models using feedback information, such as interpreter error tracebacks and database observations, to rectify any existing errors. 
This adaptive capacity is essential for agents because they have to constantly receive and react to feedback from the environment during interaction. As demonstrated in Table~\ref{tab:environment_feedback}, the performance of the \lemurSeventyChaT significantly surpasses that of the \llamaTwoSeventyChaT and \codellamaThirtyFourBInstuctioN in interactive coding benchmarks. 
This underscores the importance of having balanced capabilities for interactive agents in such environments.

We further analyze the results of InterCode-SQL to understand how models incorporate environment feedback. 
In this setting, agents are provided with database schema and guidelines for SQL game interactions. 
Acting as agents, the models are tasked with querying databases and responding to questions in a multi-turn interaction environment. 
Figure~\ref{fig:intercoder-sql} shows the Growth in Success Rate across models with each interaction turn. 
Lemur demonstrates robust performance in the first round, showing an initial performance that is comparable to that of \texttt{text-bison-001} and slightly lower than \gptturbo. 
Nevertheless, Lemur improves consistently across ten interactive rounds, surpassing the performance of \gptturbo eventually. 
In contrast, \texttt{text-bison-001}, which exhibits comparable initial performance with Lemur, does not show significant improvement. 
\llamaTwoSeventyChaT, while displaying consistent adaptability to feedback throughout the process, has a significant gap in initial performance due to inferior coding ability, hence its success rate remains relatively lower. 
\codellamaThirtyFourBInstuctioN hardly answers the questions correctly in the first round. 
We observe that this is because it blindly follows the advice in the game guide and stubbornly executes the \texttt{show tables} command first, instead of trying to understand the provided database structure. 
After an improvement in the second round, its performance returns to normal. 
However, the growth remained limited even after ten rounds of interaction, settling at par with \llamaTwoSeventyChaT and reflecting its relative weakness in adapting to environmental feedback.{\color{rebuttal}Additionally, we analyzed the probelm difficulty influence in~\S\ref{appendix:error_analysis_sql}.}

\input{tables/tab3-environmentfeedback}

\subsection{Follow Natural Language Feedback}
\label{sec:nlfeedback}
Following natural language feedback from users or other agents is an important ability for language agents. 
It requires agents to understand complex natural language instructions and convert them into symbolic executable sequences based on the current contexts. 
To evaluate the models' ability to follow natural language feedback, we follow the evaluation settings of MINT~\citep{wang2023mint}, which measures language agents' ability to leverage natural language feedback using performance improvement. 
To provide natural language feedback in multi-turn settings, MINT uses GPT-4 to simulate a user providing helpful feedback on the solutions from evaluated language agents.

We evaluate models on MINT-Reasoning and MINT-Code with and without GPT-4 feedback and the experimental results are in \Cref{tab:user_feedback}. 
MINT-Reasoning includes five modified benchmarks: GSM8k, MATH, TheoremQA, HotpotQA, and MMLU. 
MINT-Coding includes modified HumanEval and MBPP. We find that all models can benefit from GPT-4, which means powerful GPT-4 as a teacher can provide helpful feedback even without ground truth. We also calculate $\Delta_{~\texttt{feedback}}$, which indicates the absolute improvement thanks to GPT-4 feedback. 
According to the results from \Cref{tab:user_feedback}, \lemurSeventyChaT model obtain 8.19 in $\Delta_{\texttt{feedback}}$, which is significantly better than \llamaTwoSeventyChaT and \codellamaThirtyFourBInstuctioN.

\input{tables/tab4-userfeedback}

\subsection{Explore in Partially Observable Environments}
\label{sec:partiallyobservable}

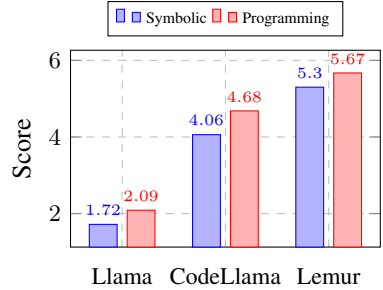
\begin{figure}[t]
\begin{minipage}{0.6\linewidth}
    \footnotesize
    \input{tables/tab5-agent}
    \label{tab:agent}
\end{minipage}
\hspace{0.15cm}
\begin{minipage}{0.35\linewidth}
        \centering
        \input{images/webarena}
        \vspace{-6.8mm}
        \captionsetup{margin={0.6cm,0cm},font=footnotesize}
        \caption{ Performance Comparison of symbolic and programming representation for WebArena.}
        \label{fig:webarena}
\end{minipage}
\end{figure}

We evaluate language agents in varied environments and find that \lemurSeventyChaT exhibits balanced and commendable performance across all tested tasks. {\color{rebuttal} As shown in \Cref{tab:agent}}, it scored $22.00$ in InterCode-CTF, showcasing its proficiency in multifaceted terminal skills and strategic adaptability. 
{\color{rebuttal} Moreover, we conducted further error analysis on CTF tasks in ~\S\ref{appendix:error_analysis_ctf} and found that besides general skills of OS terminal, CTF tasks also require cybersecurity domain knowledge. This indicates the challenging requirements for agents to perform well in different environments.}
In WebArena, it achieved a score of $5.79$, reflecting its adeptness in interpreting and executing advanced natural language commands in intricate web environments. 
In ALFWorld, it demonstrated superior planning and commonsense reasoning with a score of $59.70$, successfully navigating and performing in simulated physical environments. 
While \gptfour overall exhibits higher scores, \lemurSeventyChaT's consistent performance across diverse and partially observable environments underscores its versatility and potential in handling real-world, multifarious applications.

We also conduct experiments to explore different output formats for actions in the WebArena environment. 
Instead of merely prompting the language model to directly produce predefined actions (e.g. type [id] [content] [press\_enter\_after]), we deterministically map the action space to Python representations (e.g. type(id:int, content:str, press\_enter\_after:bool)).
We then prompt the language model to predict this representation and then parse the model's Python prediction into the predefined executable actions. 
Our findings, as shown in Figure~\ref{fig:webarena} indicate that mapping to a Python representation leads to better performance than directly predicting actions. 
Such results suggest that carefully and rationally selecting intermediate representation, based on the pre-training corpus, can effectively enhance the model's performance as a language agent, aligning with prior findings \citep{hu2022context}.

\section{Related Work}
\textbf{Transfer Learning on Code}
The advancement in expansive language models~\citep{devlinBERTPretrainingDeep2019, radfordLanguageModelsAre2019, raffelExploringLimitsTransfer2019, brown2020language} has catalyzed progress in transfer learning for code-related tasks, further enriched by the continuous pre-training paradigm~\citep{gururanganDonStopPretraining2020}. \citet{hernandezScalingLawsTransfer2021} offered insights into the interplay between model size, training data, and performance in transferring language ability to code tasks. 
Several models have been introduced, exhibiting enhanced program synthesis and infilling/completion performance, by undergoing continual pre-training on extensive code data~\citep{fengCodeBERTPreTrainedModel2020, wangCodeT5IdentifierawareUnified2021, chen2021evaluating, roziereCodeLlamaOpen2023}. 
However, their intense focus on code often results in a compromise on natural language capabilities. 
Lemur addresses this by moderately augmenting a large model with a balanced mixture of code and natural language data, maintaining proficiency in both domains.

\textbf{Instruction Fine-tuning}
The process of aligning LLMs to follow instructions, often referred to as instruction tuning, has been primarily directed towards NLP tasks~\citep{weiFinetunedLanguageModels2021, wangSuperNaturalInstructionsGeneralizationDeclarative2022}. 
Recent studies have sought to broaden the use cases of instruction tuning to involve a wider variety of general tasks~\citep{ouyang2022training}. 
Self-instruct method generates instructions using seed instructions, aiding the understanding of how to adapt language models by fine-tuning them on instructions garnered from ChatGPT~\citep{wang2022self, zhengJudgingLLMasajudgeMTBench2023, xuWizardLMEmpoweringLarge2023, mukherjeeOrcaProgressiveLearning2023}. 
We adopt a similar approach in tuning our model to follow instructions.

\textbf{Language Agents}
Language agents are adept at following user instructions and engaging with environments to execute tasks.
Recent trends in research and open-source communities have employed Large Language Models (LLMs)~\citep{brown2020language, chen2021evaluating, chowdhery2022palm, openaiGPT4TechnicalReport2023} as the principal controllers for these agents~\citep{yao2022react, LangChain, AutoGPT, shinn2023reflexion, wang2023voyager, xu2023rewoo, lin2023swiftsage, yao2023retroformer}.
This is driven by the LLMs' demonstrated abilities in reasoning, planning, grounding, and code generation~\citep{wei2022chain, huang2022language, ahn2022can, text2reward}, crucial for comprehending user instructions, grasping the environmental context, and generating executable actions.
\lemuR seamlessly integrates capabilities in both text and code, enabling the generation of environment-grounded, executable actions essential for constructing language agents.

\textbf{Agent Evaluation Benchmarks}
The rapid evolution of language agents requires an accurate and comprehensive assessment of agents. 
Recent works pose new dimensions that put language agents in web environments~\citep{deng2023mind2web, yao2022webshop, zhou2023webarena}, interactive code environments~\citep{yang2023intercode}, digital game~\citep{fan2022minedojo}, and household~\citep{puig2018virtualhome, shridhar2020alfred,shridhar2020alfworld} to finish certain task under natural language instruction.
Apart from collecting new tasks, these agent benchmarks can be established by re-purposing and transforming existing datasets to agent datasets~\citep{wang2023mint, yang2023intercode,liuAgentBenchEvaluatingLLMs2023}.
Our research combines these tasks, conducts extensive evaluation, and evaluates the model's capability to construct a language agent from multiple dimensions.

\section{Conclusion}
In conclusion, this research underscores the pivotal role of harmonizing natural and programming language proficiencies in the evolution of language models to sophisticated language agents. By developing \lemuR and \lemurChaT, we demonstrated that the meticulous amalgamation of these competencies allows for elevated performance in diverse environments and applications, narrowing the existent capability divide between open-source and proprietary models.
We open-sourced both models, intending to foster further research in the field of language models for agents.
\clearpage
\clearpage
\input{acknowledgments}
\bibliography{custom}
\bibliographystyle{iclr2024_conference}

\clearpage
\appendix
\input{appendix}

\end{document}

%% file: math_commands.tex
\usepackage{amsmath,amsfonts,bm}

\def\eqref#1{equation~\ref{#1}}

\def\1{\bm{1}}

\DeclareMathAlphabet{\mathsfit}{\encodingdefault}{\sfdefault}{m}{sl}
\SetMathAlphabet{\mathsfit}{bold}{\encodingdefault}{\sfdefault}{bx}{n}



%% file: tables/sftdata.tex
\begin{table}[t]
\centering
\setlength\tabcolsep{3pt}
\caption{Instruction datasets investigated in this work. We report the average number of rounds ($\bar{N}_{\text{rounds}}$), average length of prompts ($\bar{L}_{\text{prompt}}$), average length of completion ($\bar{L}_{\text{completion}}$). 
\label{tab:instruction-tuning-datasets} 
}
\resizebox{\textwidth}{!}{%
\begin{tabular}{llrcccc}
\toprule
\textbf{Datasets} & \textbf{Query Source} & \textbf{Response Source} & \textbf{\# Instances} & $\bar{N}_{\text{rounds}}$ & $\bar{L}_{\text{prompt}}$ & $\bar{L}_{\text{completion}}$ \\
\midrule
Open Assistant 1 & Human-written & Human-written & 34,546      & 1.6   & 34.9  & 213.1   \\
OpenOrca    & Human-written & GPT-4 & 200,000      & 1.0    & 242.8    & 174.1    \\
ShareGPT \& ChatLogs  & User prompts & GPT-3.5/GPT-4 & 81,319   & 6.0  & 96.7    & 340.2     \\
Evol-CodeAlpaca &  GPT-3.5/GPT-4     & GPT-3.5/GPT-4   & 51,952     & 1.0     & 93.9   & 343.5   \\
\bottomrule
\end{tabular}
}
\end{table}
\vspace{-10pt}

%% file: tables/agent-datasets.tex
\begin{table}[ht]
\setlength{\tabcolsep}{0.1cm}
\caption{Multi-turn agent evaluation under different settings. 
We evaluate crucial capabilities for LLM as an agent, spanning tool usage, feedback adherence and environment exploration.
}

\resizebox{\textwidth}{!}{%
\begin{tabular}{@{}lccl@{}}
\toprule
Capabilities &
  Environments &
  Observability &
  Datasets \\ \midrule
\multirow{2}{*}{Augment with Tools (\S\ref{sec:tools})} &
  Python Calculator &
  \multirow{2}{*}{Fully} &
  M-\{GSM8K, MATH, TheoremQA\} \\
 &
  WikiSearch &
   &
  M-\{HotpotQA, MMLU\} \\ \midrule
\multirow{4}{*}{\shortstack[l]{Self-debug with \\ Environment Feedback (\S\ref{sec:self-debug})}} &
  Python &
  \multirow{4}{*}{Fully} &
  M-\{HumanEval, MBPP\} \\
 &
  OS Terminal &
   &
  InterCode-Bash \\
 &
  Database &
   &
  InterCode-SQL \\
 &
  Robotics API &
   &
  RoboCodeGen \\ \midrule
\multirow{2}{*}{\shortstack[l]{Follow Natural Language \\ Feedback 
(\S\ref{sec:nlfeedback})}} &
  User/Agent &
  \multirow{2}{*}{Fully} &
  M-Reasoning w/ GPT-4 feedback\\
 &
  User/Agent &
   &
  M-Code w/ GPT-4 feedback \\ \midrule
\multirow{3}{*}{\shortstack[l]{Explore in Partially \\ Observable Environment (\S\ref{sec:partiallyobservable})}} &
  OS Terminal &
  \multirow{3}{*}{Partially} &
  InterCode-CTF \\
 &
  Web Browser &
   &
  WebArena \\
 &
 Embodied Simulator &
   &
  M-ALFWorld \\ \bottomrule
\end{tabular}%
}
\label{tab:eval_datasets}
\end{table}

%% file: tables/tab1-base-instruct.tex
\begin{table}[t]
\centering
\setlength{\tabcolsep}{0.08cm}
\sisetup{table-format=2.1}
\caption{Performance comparison across diverse models on text and code benchmarks. 
MCode is an abbreviation for Multilingual Code. 
HE stands for HumanEval. 
Avg denotes the average performance across all benchmarks. 
\lemurSeventyB and \lemurSeventyChaT exhibit balanced capabilities, achieving the highest overall performance when averaged by task.
\color{rebuttal}{See Appendix \ref{app:baseline_models} for more details.
}
}
\resizebox{\textwidth}{!}{
\begin{tabular}{
  l
  S[table-format=2.1]
  S[table-format=2.1]
  S[table-format=2.1]
  S[table-format=2.1]
  S[table-format=2.1]
  S[table-format=2.1]
  S[table-format=2.1]
  S[table-format=2.1]
  S[table-format=2.1]
}
\toprule
 & \multicolumn{3}{c}{Text}      & \multicolumn{5}{c}{Code}  & \\
\cmidrule(lr){2-4} \cmidrule(lr){5-9}
                       Model & {QA} & {Reason} & {Math}  & \multicolumn{2}{c}{Python} & {SQL}    & {MCode} & {DS} & Avg \\
\cmidrule(lr){5-6}
                       & {MMLU} & {BBH} & {GSM8K} & {HE} & {MBPP} & {Spider} & {MultiPL-E} & {DS-1000} &  \\ 
\midrule
\starcoderFiveteenB              & 30.8      & 33.2      & 8.9   & 33.6           & 52.7      & 58.3   & 25.3      & 26.0 & 33.6           \\
\starcoderPlusFiveteenB         & 42.0        & 36.2      & 17.7  & 26.2           & 37.0        & 48.8   & 21.4      & 19.4  & 31.1       \\

\codellamaThirtyFouR            & 52.8      & 42.2      & 32.7  & 48.8           & 55.0        &  68.4  & 36.4      & 31.8  & \underline{46.0}       \\

\llamaTwoSeventY            & 68.9      & 51.2      & 56.8  & 30.5           & 45.4      & 60.0     & 24.4      & 11.3   & 43.6      \\
\lemurSeventyB         & 64.5      &   51.6    & 54.9  & 35.4           & 53.2       & 62.8   &     30.4      & 30.7   & \textbf{47.9}      \\ 

\midrule

\wizardcoderFiveteenB   & 29.4      & {28.8}         & 7.1   & 57.3           &  51.6     & 61.6   & 30.8      & 29.2    & 37.0     \\

\codellamaThirtyFourBInstuctioN & 53.5      & 37.1         & 41.0    & 41.5           & 57.0      & 66.6   & 36.1      & 32.3     & \underline{45.6}    \\

\llamaTwoSeventyChaT       & 63.9      & 38.9         & 48.7  & 31.1           & 38.2      & 60.3   & 22.6      & 17.8   &   40.2    \\
\lemurSeventyChaT    &  65.3     & 61.9         &   66.3  &    61.0        &  55.5      &  62.5   &     42.9      &     34.5     &  \textbf{55.0}   \\ 
\bottomrule
\end{tabular}
}
\label{tab:base&instruct}
\end{table}

%% file: images/intercode-line.tex
\begin{tikzpicture}
\begin{semilogxaxis}[
    width=6.5cm, height=5.5cm, 
    xlabel={Turns (Log Scale)}, 
    xlabel style={yshift=1ex},
    ylabel={Performance(\%)}, 
    ylabel near ticks,
    ylabel style={xshift=-1ex},
    grid=both, 
    xmin=1, xmax=10, 
    ymin=0, ymax=80, 
    xtick={1, 2, 3, 4, 5, 6, 7, 8, 9, 10}, 
    xticklabels={1, 2, 3, 4, 5, 6, 7, 8, 9, 10}, 
    ymajorgrids=true,
    grid style=dashed,
    legend style={
        at={(1,0)},
        anchor=south east,
        xshift=-1.0mm,
        yshift=1.0mm,
        font=\scriptsize,
    },
]

    
\addplot[
    color=red,
    mark=*,
    mark size=1.5pt,thick
    ]
    coordinates {
    (1,67.99)(2,71.08)(3,72.24)(4,72.34)(5,72.73)(6,72.73)(7,72.82)(8,72.82)(9,72.82)(10,72.82)
    };
    \addlegendentry{\texttt{gpt-3.5-turbo}}

\addplot[
    color=green,
    mark=*,
    mark size=1.5pt,thick
    ]
    coordinates {
    (1,61.32)(2,70.02)(3,72.05)(4,73.31)(5,73.50)(6,73.50)(7,73.50)(8,73.60)(9,73.60)(10,73.79)
    };
    \addlegendentry{\texttt{lemur-70b-chat}}

\addplot[
    color=orange,
    mark=*,
    mark size=1.5pt,thick
    ]
    coordinates {
    (1,18.86)(2,39.65)(3,53.87)(4,58.32)(5,61.61)(6,63.54)(7,64.80)(8,65.76)(9,67.02)(10,67.89)
    };
    \addlegendentry{\texttt{llama-2-70b-chat}}

\addplot[
    color=purple,
    mark=*,
    mark size=1.5pt,thick
    ]
    coordinates {
    (1,0.77)(2,59.57)(3,64.99)(4,66.54)(5,67.12)(6,67.41)(7,67.60)(8,67.70)(9,67.79)(10,67.79)
    };
    \addlegendentry{\texttt{codellama-34b-inst}}


\addplot[
    color=pink,
    mark=*,
    mark size=1.5pt,thick
    ]
    coordinates {
    (1,61.41)(2,63.25)(3,63.64)(4,63.64)(5,63.64)(6,63.64)(7,63.64)(8,63.64)(9,63.83)(10,63.93)
    };
    \addlegendentry{\texttt{text-bison-001}}
    
\end{semilogxaxis}
\end{tikzpicture}

%% file: tables/tab2-tooluse.tex
\newcommand{\B}[1]{\multicolumn{1}{c}{\textbf{#1}}}
\vspace{10pt}
\begin{table}[ht]
\setlength{\tabcolsep}{0.1cm}
\sisetup{detect-all}
\centering
\caption{
The tool-augmented reasoning tasks evaluate the model's capabilities to use tools in the reasoning process. 
Across five tasks with Python and WikiSearch API as tools, \lemurSeventyChaT outperforms both \llamaTwoSeventyChaT and \codellamaThirtyFourBInstuctioN by large margins.
}
\resizebox{\textwidth}{!}{%
\begin{tabular}{
  @{}l
  S[table-format=2.2]
  S[table-format=2.2]
  S[table-format=2.2]
  S[table-format=2.2]
  S[table-format=2.2]
  S[table-format=2.2]
  S[table-format=2.2]
  @{}
}
\toprule
\multirow{2}{*}{Model} & \multicolumn{3}{c}{\parbox{4cm}{\centering Math Reasoning\\with Python}} & \multicolumn{2}{c}{\parbox{4cm}{\centering Question Answering \\with WikiSearch API}} & Micro Avg \\ 
\cmidrule(lr){2-6}
 & \footnotesize M-GSM8K     & \footnotesize M-MATH      & \footnotesize M-TheoremQA & \footnotesize M-HotpotQA  & \footnotesize M-MMLU                   \\ 
\midrule
\footnotesize \llamaTwoSeventyChaT      & 33.33          & 3.00           & 2.04           & 27.91          & 42.11          & 20.25        \\ 
\footnotesize \codellamaThirtyFourBInstuctioN & 25.00          & 4.00           & 2.04           & 16.28          & 30.26          & 14.87    \\ 
\footnotesize \lemurSeventyChaT      & \B{58.33} & \B{~~6.00}  & \B{~~8.16}  & \B{44.19} & \B{56.58} & \B{31.65} \\ 
\midrule
\gptturbo          & 43.75          & 26.00          & 28.57          & 27.91          & 56.58          & 36.71   \\ 
\gptfour                  & \B{93.75} & \B{57.00} & \B{57.14} & \B{46.51} & \B{80.26} & \B{66.77} \\ 
\bottomrule
\end{tabular}%
}

\label{tab:tooluse}
\end{table}

%% file: tables/tab3-environmentfeedback.tex
\begin{table}[]
\centering
\setlength{\tabcolsep}{0.1cm}
\caption{
Model performance with environment feedback. 
\lemurSeventyChaT demonstrates strong capabilities to comprehend environment feedback, achieving performance on par with \texttt{gpt-3.5-turbo}. 
*Due to overfitting to \texttt{[PYTHON]} tag, \codellamaThirtyFourBInstuctioN fails to follow feedback and generated parsable results~\citep{wang2023mint}.
}
\resizebox{\textwidth}{!}{%
\begin{tabular}{@{}lccccc@{}}
\toprule
Debug Environment & \multicolumn{2}{c}{Python Interpreter} & Bash Terminal  & Database      & Robotics    \\ \midrule
Dataset   & M-HumanEval       & M-MBPP       & IC-Bash & IC-SQL & RoboCodeGen \\ \midrule
\llamaTwoSeventyChaT       & ~~8.89$^*$  & ~~8.79$^*$  & 31.50 & 67.89 &  48.65 \\
\codellamaThirtyFourBInstuctioN & ~~2.22$^*$  & ~~2.20$^*$  & \textbf{36.00} & 67.79 &  64.86 \\
\lemurSeventyChaT      & \textbf{46.67} & \textbf{17.58} & 34.50 & \textbf{73.79} & \textbf{75.68} \\ \midrule
\gptturbo          & 37.78 & 25.27 & 46.51 & 72.82 & 83.78 \\
\gptfour           & \textbf{73.33} & \textbf{52.75} & \textbf{48.52} & \textbf{84.41} & \textbf{83.78} \\ \bottomrule
\end{tabular}%
}
\label{tab:environment_feedback}
\end{table}

%% file: tables/tab4-userfeedback.tex
\begin{table}[ht]
\centering
\vspace{10pt}
\setlength{\tabcolsep}{0.1cm}
\caption{
Comparative analysis of various models’ Successful Rate (SR) on tasks related to Reasoning and Code Generation, with and without GPT-4 feedback. 
The table presents the performance metrics of each model under two conditions: `no feedback' and `with GPT-4 feedback'. 
The Micro Avg. column represents the average performance of the models across the tasks.}

\begin{tabular}{@{}lccccc@{}}

\toprule
Model & Feedback & Reasoning & Code Gen. & Micro Avg. & $\Delta_{\texttt{feedback}}$ \\ \midrule
\multirow{2}{*}{\llamaTwoSeventyChaT}       & no feedback       & 20.25 & {8.82}  & 16.81 & \multirow{2}{*}{{5.31}}  \\
                                        & w/ GPT-4 feedback & 23.10 & 19.85 & 22.12 &                        \\ \midrule
\multirow{2}{*}{\codellamaThirtyFourBInstuctioN} & no feedback       & 14.87 & {2.21}  & 11.06 & \multirow{2}{*}{4.20}  \\
                                        & w/ GPT-4 feedback & 20.25 & 3.68  & 15.27 &                        \\ \midrule
\multirow{2}{*}{\lemurSeventyChaT}      & no feedback       & 31.65 & 27.21 & 30.31 & \multirow{2}{*}{8.19}  \\
                                        & w/ GPT-4 feedback & 35.76 & 44.86 & 38.50 &                        \\ \midrule
\multirow{2}{*}{\gptturbo}          & no feedback       & 36.71 & 29.41 & 34.51 & \multirow{2}{*}{12.39} \\
                                        & w/ GPT-4 feedback & 50.32 & 38.97 & 46.90 &                        \\ \midrule
\gptfour (upper-bound)                                   & no feedback       & 66.77 & 59.56 & 69.11 & {N/A}          \\ \bottomrule
\end{tabular}%
\label{tab:user_feedback}
\end{table}

%% file: tables/tab5-agent.tex
\setlength{\tabcolsep}{0.1cm}
\captionof{table}{Performance comparison of different models in partially observable environments InterCode-CTF, WebArena and ALFWorld.}
\vspace{1.0mm}
\begin{tabular}{@{}lcccc@{}}
\toprule
\multirow{2}{*}{Model} & \multicolumn{2}{c}{Digital Env.}     & Physical Env. \\ 
\cmidrule(l){2-4} 
                       & IC-CTF        & WebAreana            & ALFWorld             \\ \midrule
\footnotesize \llamaTwoSeventyChaT       & 9.00                 &  ~~1.72                 & 21.64                \\
\footnotesize \codellamaThirtyFourBInstuctioN & 16.00                & ~~4.06               & 37.31                \\
\footnotesize \lemurSeventyChaT      & \textbf{22.00}                & \textbf{~~5.30}                 & \textbf{59.70}                \\ \midrule
\footnotesize \gptturbo          & 11.00                & ~~7.38               & 41.79                \\
\footnotesize \gptfour                  &\textbf{37.00}                & \textbf{10.59}                & \textbf{84.33}  
 \\ \bottomrule
\end{tabular}
\label{tab:partiallyobservable}

%% file: images/webarena.tex
\begin{tikzpicture}
    \begin{axis}[
            ybar=3.8pt,
            enlargelimits=0.15,
            enlarge x limits=0.25,
            width=5.7cm, height=4.2cm,
            bar width=0.37cm,
            grid=both,
            grid style=dashed,
            every axis label={font=\footnotesize},
            tick label style={font=\footnotesize},
            legend style={at={(0.5,1.25), font=\tiny},
            anchor=north,legend columns=-1},
            xlabel style={font=\footnotesize},
            ylabel={Score},
            ylabel style={font=\footnotesize,xshift=2.5em},
            ylabel near ticks,
            symbolic x coords={Llama, CodeLlama, Lemur},
            xtick=data,
            xlabel near ticks,
            xtick style={draw=none},
            nodes near coords,
            nodes near coords align={vertical},
            nodes near coords style={font=\tiny}
        ]
        \addplot coordinates {(Llama,1.72) (CodeLlama,4.06) (Lemur,5.3)};
        \addplot coordinates {(Llama,2.09) (CodeLlama,4.68) (Lemur,5.67)};
        \legend{Symbolic,Programming},

    \end{axis}
\end{tikzpicture}

%% file: acknowledgments.tex
\section{Acknowledgments}
This project is an open research collaboration between XLang Lab of HKU and Salesforce Research. We extend our heartfelt gratitude for the generous research grants received from both Google Research and Amazon AWS. Our special thanks go to the Google TPU Research Cloud program for their provision of vital computational resources. Our research greatly benefitted from insightful discussions with esteemed collaborators: Pengcheng Yin, Yaqing Wang, and You Wu from Google Research; Peng Shi and Shuaichen Chang from AWS AI; Yizhong Wang from the University of Washington; Xinyang Geng from UC Berkeley; Ansong Ni from Yale University; Chen Henry Wu from Carnegie Mellon University; Zeyu Liu from the University of Texas at Austin; Lin Zheng and Jiacheng Ye from the University of Hong Kong.

%% file: appendix.tex
\section{Lemur and Lemur-Chat}
\subsection{Pre-training}
\label{appendix:pt}
\subsubsection{Pre-training Data}
\label{appendix:ptdata}

We performed pre-training on the top of \llamaTwO. The detailed statistics of pre-training data corpus is presented at below in Table~\ref{tab:ptdata}.
\vspace{2mm}
\input{tables/ptdata}

{
\color{rebuttal}
\subsubsection{Data Mixture Ratio Discussion}
\label{appendix:ratio}
The data mixture for continual pre-training is a complex trade-off between the model size, the vanilla performance of the model, data quality, learning efficiency, catastrophic forgetting etc~\cite{hernandezScalingLawsTransfer2021}.
To roughly estimate a good ratio of composing the training corpora, we conducted a continue pre-training study with Llama models on a set of different code-text ratios and found that the ratio of 10:1 is an efficient ratio for the Llama model to transfer from text to text-code balance.

Due to computational limits, it is difficult for us to conduct comprehensive experiments and perform systematic studies on this scale. However, we hope that our open-source effective settings and training checkpoints can benefit the community for continual exploration.
We believe that a method to predict the optimal continue-pre-training data mixture ratio for a pair of domains to maximize their performance would be very meaningful and interesting, and it is still an open research question for the current large language model~\cite{hernandezScalingLawsTransfer2021, chowdhery2022palm, hoffmannTrainingComputeOptimalLarge2022, aghajanyan2023scaling}.

When generalizing this approach to different models, the data mixture ratio and training steps need further adjustment. For example, smaller models have less capacity for harmonizing both text and code capabilities. Therefore, they may suffer from relatively obvious forgetting of natural language knowledge.. The strategy of data mixture for large language model pretraining is an open and valuable research question. We believe we will have a more efficient and predictable data mixture strategy in future studies.
}



\subsubsection{Training Details}
\label{appendix:pttraining}
We train our model on a TPUv4-512 pod. Our codebase is based on Jax and EasyLM~\citep{geng2023easylm}. Following the pretraining methodology of Llama 2 \citep{touvronLlamaOpenFoundation2023}, we used a batch size of 4M tokens. To improve training efficiency, we packed multiple shorter sequences into each batch entry when possible, an approach known as sequential packing~\citep{raffelExploringLimitsTransfer2019}.

Optimization was performed with Adam using a peak learning rate of 4e-5 along with $\beta_1$ = 0.9 and $\beta_2$ = 0.95. Gradients were clipped at 1.0 to prevent exploding gradients. A cosine decay schedule was used for the learning rate, with a linear warmup of 2000 steps followed by decaying the learning rate to 10.\% of its peak value at the end of training.

\subsection{Instruction Fine-tuning}
\label{appdx:ift}

\subsubsection{Instruction Fine-tuning Data}

\paragraph{OpenORCA} The OpenOrca dataset is a collection of augmented FLAN Collection data. Currently ~1M GPT-4 completions, and ~3.2M GPT-3.5 completions. It is tabularized in alignment with the distributions presented in the ORCA paper~\citep{OpenOrca} and currently represents a partial completion of the full intended dataset, with ongoing generation to expand its scope. The data is primarily used for training and evaluation in the field of natural language processing.

\paragraph{OpenAssistant 1} OpenAssistant(OASST1) is a crowdsourced human-annotated assistant-style conversation corpus comprising 161,443 messages, enriched with 461,292 quality ratings.~\citep{kopfOpenAssistantConversationsDemocratizing2023}. The data results from the collaboration of over 13,500 volunteers worldwide.

\paragraph{ShareGPT and Chatlogs} We curated English human instructions from ShareGPT and Chatlogs for instruction-tuning. To filter English instructions, we utilized the langdetect package in Python and eliminated any instructions with non-English detection results. Additionally, considering that non-English instructions often contain consecutive non-English characters, we implemented a second filtering step, removing all sentences with three or more consecutive non-English characters. To ensure semantic diversity, we employed instructor\citep{su2023embedder} to encode the filtered instructions, calculate cosine similarity, and remove instructions with a similarity score greater than 0.95. After deduplication, we obtained nearly 80K instances, with an average of about 6 rounds of high-quality data per instance.

\paragraph{Evol-CodeAlpaca} We use two open-sourced Evolution-Instruct~\citep{luo2023wizardcoder} datasets, i.e., ~\cite{evol80kdata} and~\cite{codealpaca}, and an execution-verified Python dataset constructed by us. After applying the same deduplication method as the Text data, we obtained $\sim$ 46K examples. 

\subsubsection{Instruction Fine-tuning Details}
We use Huggingface Transformers~\citep{Wolf2019HuggingFacesTS} with Accelerate Library to fine-tune the Lemur model to obtain the Lemur-Chat model. We train on our data collection for two epochs. We use Adam optimizer with a learning rate of 2e-5 and a batch size of 128.

\section{Agent Evaluation}
\label{appendix:agent-eval}
\textbf{MINT}~\citep{wang2023mint} is a well-rounded evaluation that covers a range of tasks repurposed for multi-turn evaluation. 
It consists of three types of tasks, namely reasoning (MMLU~\citep{hendrycks2021measuring}, GSM8K~\citep{cobbe2021training}, MATH~\citep{hendrycks2021measuring}, TheoremQA~\citep{chen2023theoremqa}, HotpotQA~\citep{yang2018hotpotqa}), code generation (HumanEval~\citep{chen2021evaluating}, MBPP~\citep{austin2021program}), and decision-making (ALFWorld~\citep{shridhar2020alfworld}). 
To assess the proficiency of language models in employing tools, their reasoning process is reoriented to incorporate tool use. 
For instance, language models are prompted to utilize the Python calculator to work out mathematical problems, rather than supplying the answer outright.
In code generation, the LLM is encouraged to incorporate Python interpreter messages to check generated code.
To prevent any misunderstanding, we use the prefix ``M-'' to differentiate the original dataset from the MINT version in our paper. 

\textbf{InterCode-Bash/SQL}~\citep{yang2023intercode} are two tasks that serve as an experimental platform that assesses the capacity of extensive language models to integrate feedback from the environment during interactive coding.
This benchmark evaluates models in a way of generating a series of actions under user instruction, and regards elements such as execution outcomes, and error backtracking, amongst others as environment observations.

\textbf{RoboCodeGen}~\citep{liang2023code} serves as a specialized evaluation framework focused on robotics-related tasks.
It comprises three main types of questions that target spatial reasoning (e.g., identifying the closest point among a set of points), geometric reasoning (e.g., verifying if one bounding box is contained within another), and controls (e.g., PD control).

\textbf{InterCode-CTF} is a task in the InterCode evaluation suite. 
Capture the Flag (CTF) is a competitive cybersecurity game that requires LLMs to discover encrypted ``flag'' hidden within code snippets or file systems.
Compared with Bash and SQL generation, CTF is much more complex, requiring agents to have knowledge of multiple coding languages, modularize higher-order objectives into sub-problems, create multi-step plans for solving each problem, and adjust strategies when a plan fails to provide any useful insights.

\textbf{WebArena}~\citep{zhou2023webarena}
creates self-hostable websites of four popular categories by simulating real-world equivalent functionalities and data. 
To simulate human problem-solving abilities, WebArena also embeds tools and knowledge resources as standalone websites. 
\textbf{ALFWorld}~\citep{shridhar2020alfworld} is a synthetic environment benchmark adapted from Alfred~\citep{shridhar2020alfred} in the text-based interface where agents need to navigate in simulated households (e.g., go to coffee table 1, pick up paper 2, use desk lamp 1) and achieve high-level goals (e.g., check the paper under the desk lamp).
Task instances may involve over 50 locations and 50 steps to solve, thus challenging the agent's ability to plan, track sub-goals, and conduct systematic exploration.

\color{rebuttal}{
\section{Baseline models}
\label{app:baseline_models}
\textbf{\starcoderFiveteenB}~\citet{liStarCoderMaySource2023} is a model with 15.5B parameters and 8k context length. 
It is first trained on 1 trillion tokens sourced from The Stack \cite{kocetkovStackTBPermissively2022}, a large collection of permissively licensed GitHub repositories, and then finetuned on 35B Python tokens.

\textbf{\starcoderPlusFiveteenB}~\citet{liStarCoderMaySource2023} is similar to StarCoder-15B, except that it is finetuned in RefinedWeb~\citet{penedoRefinedWebDatasetFalcon2023}, The Stack \cite{kocetkovStackTBPermissively2022} and Wikipedia dataset.
It is expected to have more balanced text and code capabilities.

\textbf{\llamaTwO}~\citet{touvron2023llama} is a model with 70B parameters trained on 2 trillion tokens from public sources.
Its training corpus is mostly natural language texts, which enables its strong abilities in textual understanding.

\textbf{\llamaTwoSeventyChaT}~\citet{touvron2023llama} finetunes Llama-2-70B and is optimized for dialogue use cases.

\textbf{\codellamaThirtyFouR}~\cite{roziere2023code} finetunes Llama-2-34B on code-intensive corpus with 500B tokens.

\textbf{\codellamaThirtyFourBInstuctioN}~\cite{roziere2023code}, Code Llama - Instruct is based on Code Llama and fine-tuned
with an additional approx. 5B tokens to better follow human instructions.

\textbf{\wizardcoderFiveteenB}~\citet{luo2023wizardcoder} was trained with instrucions on code-intensive datasets.
It achieved superior performance on popular code generation benchmarks including HumanEval, HumanEval+, MBPP, and DS1000.

\section{Error Analysis on Language Agent Tasks}
\label{appendix:error_analysis}
\subsection{Augment with Tools: A Case Study on MINT-MATH}
\label{appendix:error_analysis_math}
\input{tables/error_analysis_math}
In Table~\ref{app:error_analysis_math}, we report the percentage of examples in MINT-MATH dataset with execution errors and invalid actions.
From the results, we can see that \llamaTwoSeventyChaT produces the most execution error, followed by \codellamaThirtyFourBInstuctioN, \lemurSeventyChaT, \gptturbo and \gptfour has the fewest error.
This is aligned with the performance of task success rates reported in Table~\ref{tab:tooluse}, which indicates that execution error is an important factor in explaining the performance difference between models.

On the other hand, the rate of invalid actions among different models also follows similar trends, except that \llamaTwoSeventyChaT achieves extremely low error rate.
With further investigation into the example output, we find out that \llamaTwoSeventyChaT usually writes trivial and ineffective actions, which does not contribute to solving the problem.

\subsection{Self-debug with Environment Feedback: A Case Study on InterCode-SQL}
\label{appendix:error_analysis_sql}
\input{tables/error_analysis_sql}
In Table~\ref{app:error_analysis_sql}, We report detailed scores of different models in various spectrums of the Spider evaluation set divided by difficulty level.
According to Table~\ref{app:error_analysis_sql}, as the difficulty level increases, the performance gap between \gptfour and \lemurChaT gradually widens.
When evaluated with easy questions, the model usually solves the problem in the first round; while in the difficult split, the model needs to iteratively incorporate the environment feedback.
The larger gap between \lemurChaT and \gptfour indicates better abilities of \gptfour to perform multi-turn interaction with the environment.

\subsection{Explore in Partially Observable Environment : A Case Study on CTF}
\label{appendix:error_analysis_ctf}
\begin{figure}[ht]
    \centering
    
    \includegraphics[width=0.95\textwidth]{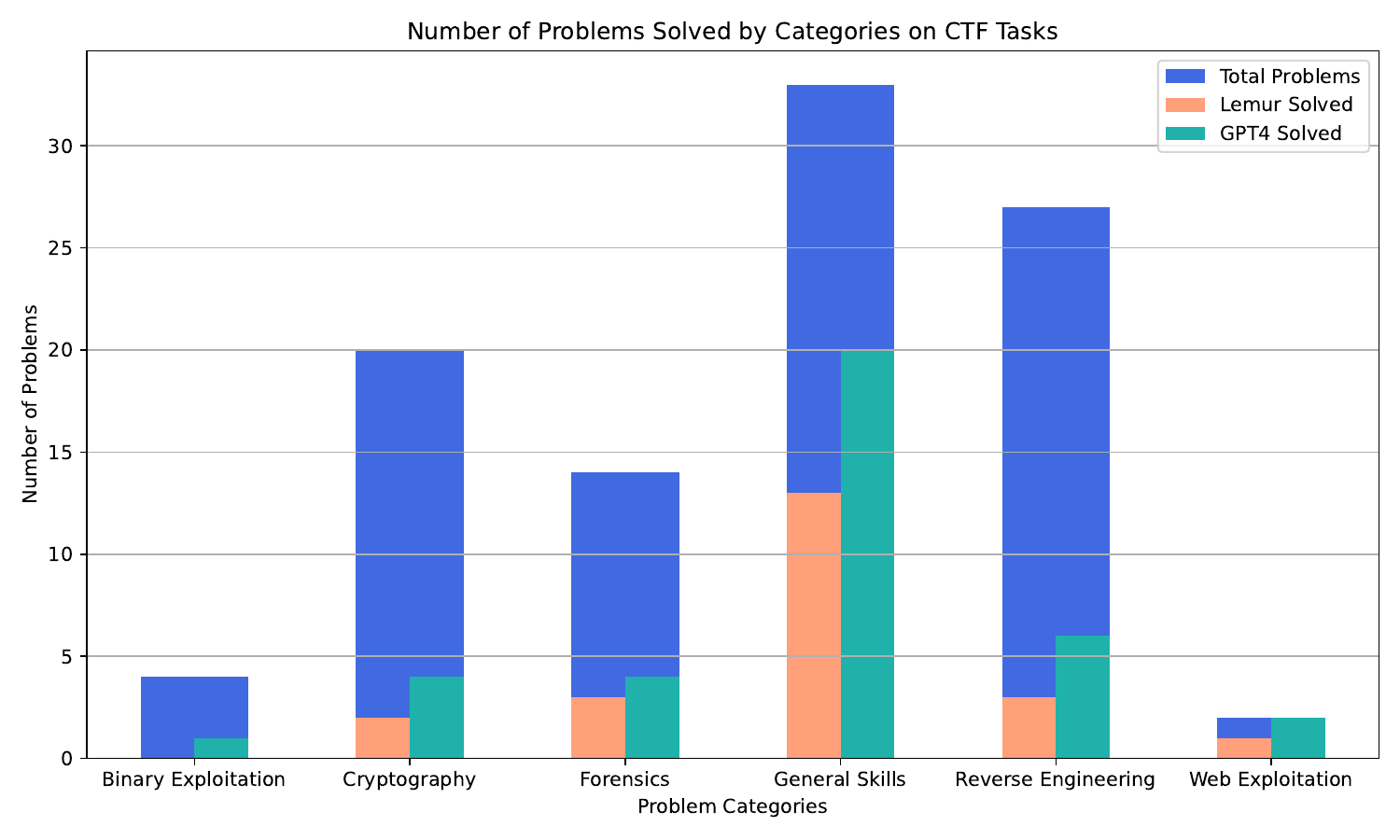}
    \caption{\color{rebuttal} The number of problems solved by \lemurSeventyChaT and \gptfour. The blue part indicates the total number of problems.}
    \label{fig:ctf_details}
\end{figure}

We manually researched the CTF (Capture The Flag) tasks, which is a popular competition program originating from cybersecurity. We manually labeled the problems in 100 CTF tasks and divided each problem into 6 categories. Figure~\ref{fig:ctf_details} shows the performance comparison between \lemurSeventyChaT and \gptfour. We found that \gptfour and \lemurSeventyChaT perform significantly better in the "general skills" category than in domain-specific fields such as "cryptography" and "reverse engineering". This means current language agents methods easily fail on domain-specific scenarios, which guide the future research of language agents.

}

%% file: tables/ptdata.tex
\begin{table}[htbp]
    \centering
    \caption{Distribution of data by Type and Source, as well as the sampling weights. Epochs are the number of passes over each constituent dataset during a full epoch over the data split.
    }
    \resizebox{\textwidth}{!}{%
    \begin{tabular}{l c | l S[table-format=2.2] S[table-format=2.2] S[table-format=1.2]}
        \toprule
        \textbf{Type} & \textbf{Weights (\%)} & \textbf{Source} & {\textbf{Weights (\%)}} & {\textbf{Effective Tokens (B)}} & {\textbf{Epoch}} \\
        \midrule
        \multirow{10}{*}{Code} & \multirow{10}{*}{90.9} & Python         & 72.73 & 65.46 & 2.98 \\
                               & & SQL            &  5.15 &  4.63 & 0.69 \\
                               & & Java           &  1.82 &  1.64 & 0.06 \\
                               & & Shell          &  1.82 &  1.64 & 1.26 \\
                               & & Notebook       &  1.71 &  1.54 & 0.82 \\
                               & & JavaScript     &  1.69 &  1.52 & 0.06 \\
                               & & C              &  1.61 &  1.44 & 0.06 \\
                               & & PHP            &  1.33 &  1.19 & 0.06 \\
                               & & CPP            &  1.21 &  1.09 & 0.06 \\
                               & & Others         &  1.73 &  1.55 & 0.06 \\
        \midrule
        \multirow{6}{*}{Text} & \multirow{6}{*}{9.1} & RefinedWeb     &  6.82 &  6.14 & {--} \\
                               & & Wikipedia      &  0.64 &  0.57 & {--} \\
                               & & Books          &  0.41 &  0.37 & {--} \\
                               & & ArXiv          &  0.41 &  0.37 & {--} \\
                               & & StackExchange  &  0.41 &  0.37 & {--} \\
                               & & DM Mathematics &  0.41 &  0.37 & {--} \\
       \midrule
       Total & 100 & & 100.00 & 90.00 & \\
        \bottomrule
    \end{tabular}
    \label{tab:ptdata}
    }
\end{table}

%% file: tables/error_analysis_math.tex
\begin{table}[ht]
\color{rebuttal}
\centering
\vspace{10pt}
\setlength{\tabcolsep}{0.1cm}
\label{app:error_analysis_sql}
\caption{\color{rebuttal}Model error rate in MATH evaluation set by causes
} 
\begin{tabular}{@{}lcc@{}}

\toprule
Model & Execution error & Invalid action \\ \midrule

\llamaTwoSeventyChaT  & 40.00 &  8.00         \\ \midrule
\codellamaThirtyFourBInstuctioN  &  38.00 &   33.00        \\ \midrule
\lemurSeventyChaT  & 29.00 &    26.00       \\ \midrule
\gptturbo  & 17.00 &  15.00   \\ \midrule
\gptfour  & 7.00 &  2.00   \\ 

\bottomrule
\end{tabular}%
\label{app:error_analysis_math}
\end{table}

%% file: tables/error_analysis_sql.tex
\begin{table}[ht]
\color{rebuttal}
\centering
\vspace{10pt}
\setlength{\tabcolsep}{0.1cm}
\label{app:error_analysis_sql}
\caption{\color{rebuttal}Model accuracy in different spectrum of Spider evaluation set
} 
\begin{tabular}{@{}lccccc@{}}

\toprule
Model & Easy & Medium & Hard & Extra & All \\ \midrule

\llamaTwoSeventyChaT  & 89.92  & 69.73 & 59.77 & 38.55 &  67.89         \\ \midrule
\codellamaThirtyFourBInstuctioN  &  90.73  & 67.49 & 63.79 & 38.55 &   67.80        \\ \midrule
\lemurSeventyChaT  & 92.74  & 74.89 & 70.69 & 45.78 &    73.79       \\ \midrule
\gptturbo  & 92.74 & 74.89 & 67.24 & 43.37 &  72.82   \\ \midrule
\gptfour  & 95.16  & 82.96 & 86.21 & 68.67 &  84.14    \\ 

\bottomrule
\end{tabular}%
\label{app:error_analysis_sql}
\end{table}